\documentclass[10pt,letterpaper]{article}
\usepackage[top=0.85in,left=2.75in,footskip=0.75in]{geometry}

\usepackage{todonotes}

\usepackage{amsmath,amssymb}

\usepackage{changepage}

\usepackage[utf8x]{inputenc}

\usepackage{textcomp,marvosym}

\usepackage{cite}

\usepackage{nameref,hyperref}

\usepackage{booktabs}

\usepackage[right]{lineno}

\usepackage{microtype}
\DisableLigatures[f]{encoding = *, family = * }

\usepackage{array}

\newcolumntype{+}{!{\vrule width 2pt}}

\newlength\savedwidth

\raggedright
\usepackage[aboveskip=1pt,labelfont=bf,labelsep=period,justification=raggedright,singlelinecheck=off]{caption}

\makeatletter
\renewcommand{\@biblabel}[1]{\quad#1.}
\makeatother

\usepackage{lastpage,fancyhdr,graphicx}
\usepackage{epstopdf}
\fancyheadoffset[L]{2.25in}
\fancyfootoffset[L]{2.25in}
\usepackage[normalem]{ulem}

\begin{document}
\vspace*{0.2in}

\begin{flushleft}

{\Large

\textbf{A deep language model to predict metabolic network equilibria} 

}
François Charton\textsuperscript{1,\Yinyang}
Amaury Hayat\textsuperscript{2\Yinyang},
Sean T. McQuade\textsuperscript{3},
Nathaniel J. Merrill\textsuperscript{4},
Benedetto Piccoli\textsuperscript{3},
\\
\bigskip
\textbf{1} Facebook AI Research 
\\
\textbf{2} CERMICS, Ecole des Ponts ParisTech, Champs-sur-Marne, France
\\
\textbf{3} Department of Mathematical Sciences and Center for Computational and Integrative Biology, Rutgers University–Camden, 303 Cooper St, Camden, NJ, USA
\\
\textbf{4} Pacific Northwest National Laboratory, Richland, WA, USA
\\
\bigskip

%
%
\Yinyang These authors contributed equally to this work.






\end{flushleft}
\section*{Abstract}
We show that deep learning models, and especially architectures like the Transformer, originally intended for natural language, can be trained on randomly generated datasets to predict to very high accuracy both the qualitative and quantitative features of metabolic networks. Using standard mathematical techniques, we create large sets (40 million elements) of random networks that can be used to train our models. These trained models can predict network equilibrium on random graphs in more than $99\%$ of cases. They can also generalize to graphs with different structure than those encountered at training. Finally, they can predict almost perfectly the equilibria of a small set of known biological networks. Our approach is both very economical in experimental data and uses only small and shallow deep-learning model, far from the large architectures commonly used in machine translation.
Such results pave the way for larger use of deep learning models for problems related to biological networks in key areas such as quantitative systems pharmacology, systems biology, and synthetic biology.


\section*{Author summary}

Many important biological processes can be modelled as metabolic networks: directed graphs with chemical elements or metabolites flowing through the edges and processed in the nodes. For a given network, finding whether an equilibrium exists, and the concentrations at every node, are important tasks in computational biology.
We show that a category of deep neural networks, primarily used for natural language translation, can be trained on synthetically generated graphs to predict with high accuracy the existence of an equilibrium and the concentrations at every node. Models trained on randomly generated graphs achieve high accuracy on real biological data, and on networks with different properties that those from the training set. We discuss the representations and encoding we use for the problems and solutions and the architectural decisions in designing the models.
Our approach constitutes a proof of concept and a first step towards using natural language models, such as the transformer architecture, for problems of computational biology. It also provides an economical way to discover and study metabolic graphs. We believe such techniques can be extended to other areas of science, and different computational problems.

\section{Introduction}
Computational methods are increasingly accepted 
and used in biomedical sciences, also 
as substitutes for costly experimental
activities and clinical trials.
For instance,
in the broad area of drug discovery  
computational methods are used for many purposes
ranging from identifying novel targets all the way to virtual patient populations
for expected clinical response. Comprehensive and system level models are the core part of Quantitative Systems Pharmacology (briefly QSP)
which has been applied to a number of therapeutic areas
 \cite{QSPSanofi2020,BR18,Cosbi20,Kaddi2018}.
Because of this, there is an increasing attention on the
impact that such methods may have in the mid and long-term horizon,
in terms of accelerating discovery and facilitating development,
\cite{GADKAR16,Nijsen2018,NIH11}.
At the same time, the advancement in data availability and
data science techniques allows for a number of large-scale projects,
such as the Physiome Project, Biomodels, and the Human Cell Atlas Model,
see \cite{QSPH21} and references therein.

In this paper, we provide a proof of concept that AI tools can be trained to efficiently substitute well-designed but computationally costly
and, often times, not scalable computational methods.
In order to achieve this goal, we focus on metabolic networks
and the characterization of network equilibria.
Many QSP models, including PBPK \cite{PBPK16},
are based on metabolic representation of drug absorption and its effect on the systems of interest. The model consists of a graph (mathematical
term for network) with  nodes representing compounds and edges
representing fluxes of the network (e.g. chemical reactions). 
A wide literature addressed the problem of characterizing
the properties of these networks, in particular
for discovering correlation among fluxes
\cite{caughman2006,feinberg1974,jacquez1993,maeda1978,mirzaev2013,palsson2015}.
The existence and characterization of equilibria is a necessary
condition to use efficiently such models \cite{LIFE2017}.
Therefore, we consider the characterization of networks
admitting equilibria for fixed flux levels and its computation.
More precisely, we focus on two specific problems:\\
(1) Given a general metabolic network with inflows and outflows,
is it possible to have an equilibrium?\\
(2) If (1) is answered positively, can we compute such equilibrium (assuming uniqueness)?\\
The answer to the first question is related to the network topology: if all nodes connected to inflows are also connected to outflows, then there exists an equilibrium \cite[Section 4.1.2]{merrill2019stability} .
Because of this characterization, it is possible to write codes
for determining the existence of equilibria and thus generate data
to train neural networks to identify metabolic networks having
such property.
The second problem can be solved explicitely for networks
with linear dynamics by a matrix inversion, see \cite{merrill2019stability,NEWLIFE},
as $J^{-1}(f)\varphi$, where $J(f)$ is the Jacobian matrix
of the linear dynamic in terms of the network fluxes $f$
and $\varphi$ is the influx vector.
Similarly to the first problem, we can easily generate codes
to compute the equilibria and train neural networks.

In this paper we show that transformers, deep neural architectures originally designed for machine translation, can be trained on randomly generated graphs to predict both the existence and the numerical values of an equilibrium to very high accuracy. 

Even though we already have algorithms for these problems, we believe that demonstrating that they can be learned by transformers is important if we want to use deep neural networks to solve problems of computational biology. It shows that transformers can be used as end-to-end solutions, without needing to ``step out" and call an external algorithm every time a computation has to be done. We believe such an approach could be extended to more difficult problems and even problems that we are not yet able to solve.

Our approach has two main advantages. First, it is very economical: since all training is performed on synthetic datasets, there is no need to collect large amounts of biological data. Besides, our models are much smaller than the very large architectures developed for natural language (e.g. GPT-3 \cite{brown2020language}), they can be trained with limited computing resources and once the model is trained, prediction of the properties of a specific biological graph is fast, easily parallelizable, and need little computing resources. This makes our approach affordable for research groups that do not have access to massive clusters and contributes to the democratization of AI as an effective tool for computational biologists. 
Second, it has very general applicability. We show that, as long as random sets of problems and solutions can be generated, small language models can accurately compute the numerical solutions of biological problems from their symbolic representations. This could potentially be applied to a wide range of problems of computational biology, even outside QSP.

\section{Results}
\label{sec:results}
\subsection{Metabolic networks and their equilibria}
\begin{figure*}[h!]
\begin{center}
\includegraphics[width=0.95\textwidth]{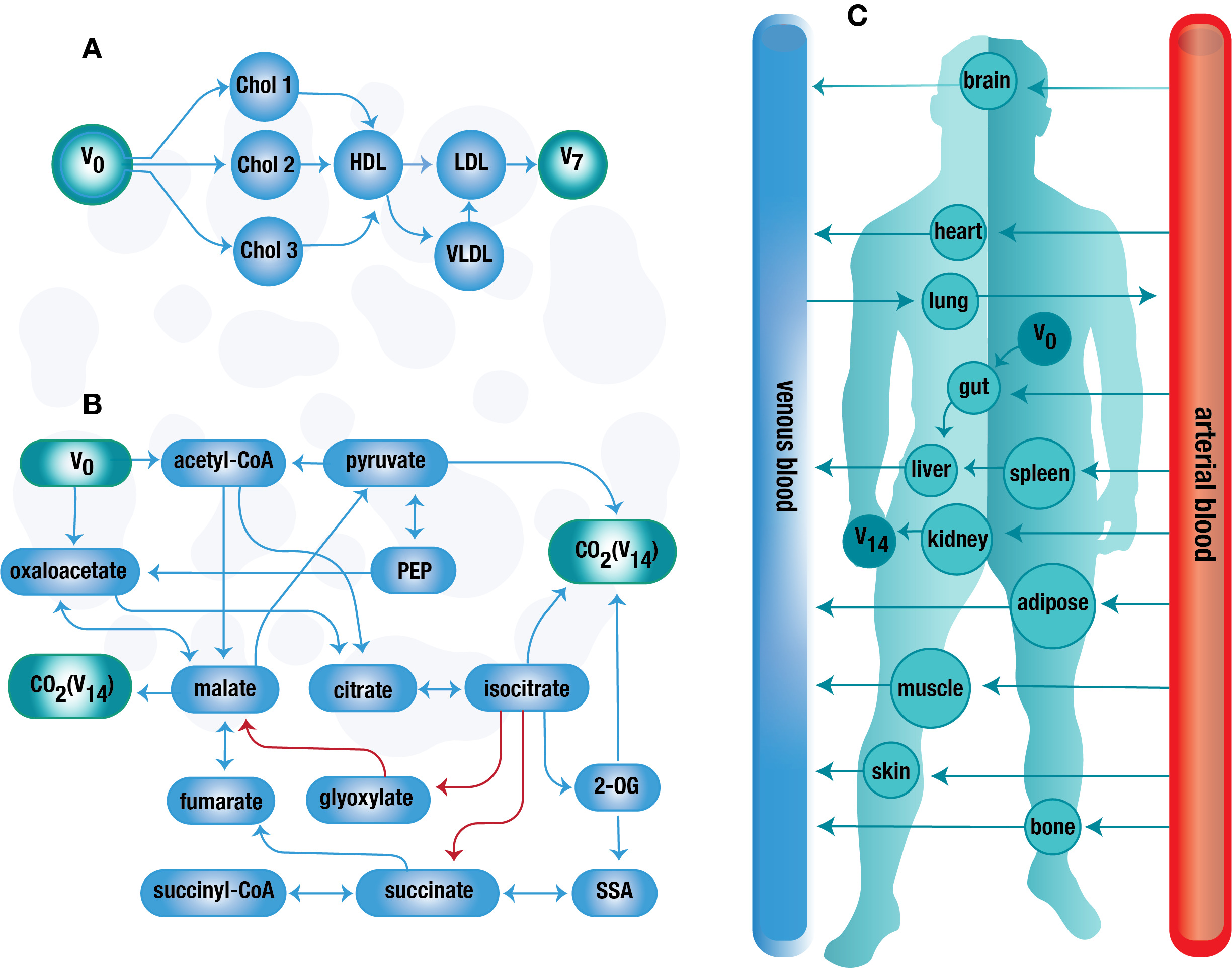}
\caption{\textbf{Three examples of metabolic graphs.} Nodes $v_0$ and $v_{n+1}$ represent the source and sink.
\textbf{A Reverse cholesterol transport \cite{LIFE2017}:} removing cholesterol from plaque build up in arteries. High Density Lipoproteins (HDL, or ``good cholesterol'') sequester and transport cholesterol from plaque, and deliver it to the liver that safely removes it from the body. 
\textbf{B Tricarboxilic acid cycle in an engineered cyanobacteria \cite{durall2021production}.} Carbon metabolism of a modified strain of the cyanobacteria Synechocystis, engineered to include a glyoxylate shunt allowing it to produce succinate, an important industrial chemical. Engineered paths are in red. 
\textbf{C PBPK model \cite{jones2009modelling}:} Tracking medication as it passes from one organ or tissue to the next. Medication ingested orally moves into the gut, but will be used to treat a different tissue of the body.\label{fig:graph_examples}}
\end{center}

\end{figure*}

A general metabolic network can be represented as a directed graph. Its nodes represent metabolites, and its edges biological interactions, for instance, chemical reactions linking reactants to products. To account for differences in the strength of interactions (e.g. the stoechiometry of chemical reactions), edges are usually weighted.
Interactions between the network and its environment consist of intakes of molecules needed by the metabolism (e.g. ingesting food) and excretion of waste material. Without loss of generality, all intakes and excretions can be represented as a pair of virtual nodes (source and sink) connected to the graph.
This representation of networks as directed graphs is common in many fields of research (e.g. transportation, telecommunications, gas and power-grid networks, social interactions). We focus here on biological networks like those presented in Figure~\ref{fig:graph_examples}.

\begin{figure}[h!]
    \centering
    \includegraphics[scale=0.6]{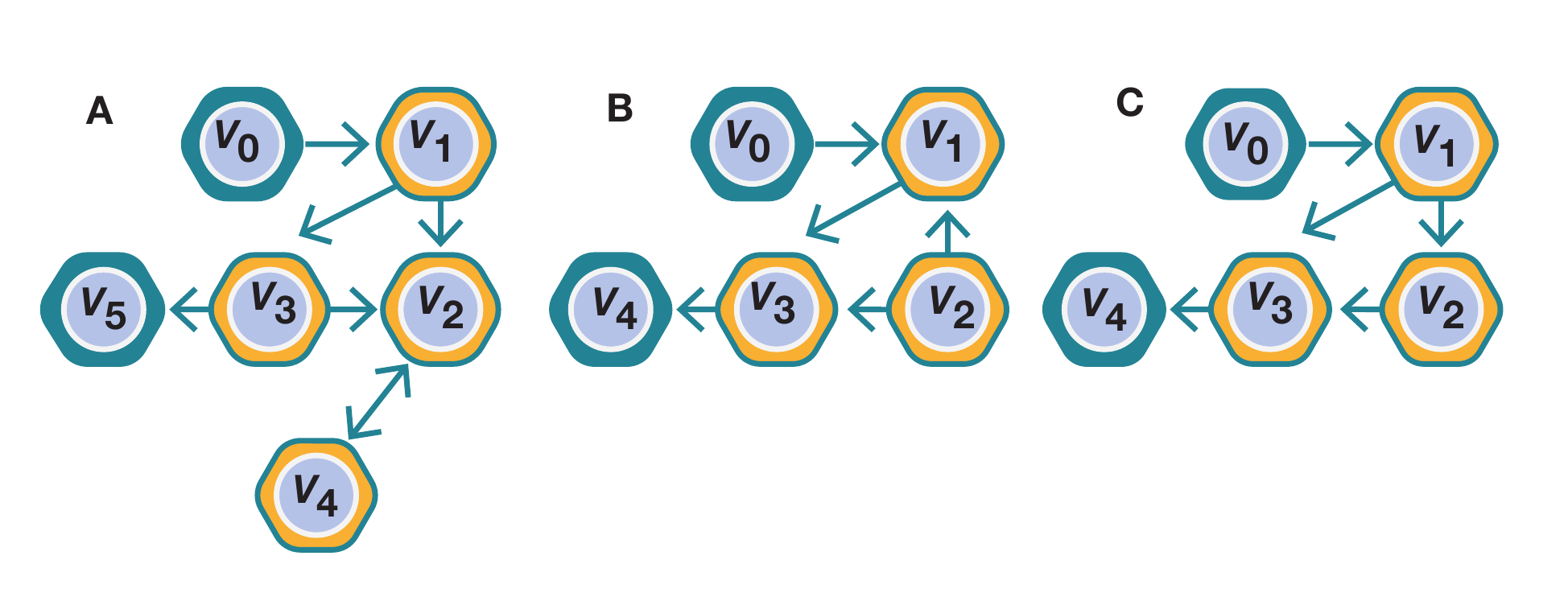}
    \caption{\textbf{Examples of networks with and without equilibrium.} \textbf{A} has no equilibrium because the nodes $v_{2}$ and $v_{4}$ have no path to excretion. \textbf{B} has no positive equilibrium because $v_2$ is not linked to intake $v_0$.    \textbf{C} has an equilibrium because all nodes linked to the intake $v_{0}$ have a path to excretion $v_{4}$.\label{fig:ex_with_eq}}
\end{figure}

For a directed graph to represent a metabolic network, it must make biological sense. First and foremost, it must have an equilibrium: a steady state of the underlying dynamic, defined by the concentrations of metabolites in each node. 
This can be shown to be equivalent with having each node connected to intakes also connected to excretions, so that metabolites flow through the graph, and there are no ``dead-ends'' where they accumulates indefinitely \cite{merrill2019stability}. Examples of graphs with and without equilibrium are given in Figure \ref{fig:ex_with_eq}.

We consider two problems: determining whether a graph has an equilibrium (the qualitative problem), and, for a graph with an equilibrium (and a dynamic linear with respect to metabolites), determining the concentrations of metabolites in each node for a given value of intakes (the quantitative problem). Numerical methods have been proposed that solve these problems 
by leveraging the link between the adjacency matrix of the directed graph and the dynamic of the metabolites, reformulating both problems as computations over graphs and adjacency matrices (\cite{merrill2019stability} and Supplementary~\ref{appendix:maths}). Specifically, the qualitative problem can be 
solved by finding paths between every internal node connected to the intake and the virtual excretion node, and the quantitative problem can be solved by removing from the adjacency matrix the rows and columns corresponding to intakes, creating a diagonal matrix from the row sums of this reduced matrix, and substracting it from the reduced matrix. Removing from the resulting matrix the rows and columns corresponding to the excretions gives a so-called grounded Laplacian. Taking the transpose of the grounded Laplacian and multiplying its inverse by the vector of intakes yields the concentrations at equilibrium. 

In this paper, we show that these two problems can be solved, instead, by a deep language model.

\subsection{Using deep learning models to predict equilibria}
Designing an AI model usually consists of three steps: defining a representation for the problem and an architecture for the model, training it on a dataset of relevant examples, and evaluating its accuracy on smaller test sets of held-out examples not seen at training, or on real-life data. 
Here, the limiting factor is the training data: deep learning models typically require datasets of millions of examples, much more than could be collected from known metabolic networks. We address this issue by creating synthetic datasets of randoms graphs to train our models. Using an Erd\H{o}s-R\'enyi model \cite{gilbert1959random,erdHos1960evolution}, we generate graphs  by randomly selecting a number of nodes and edges ($n$ and $e$, with $2n \leq e \leq 4n$), randomly selecting each of the $n(n-1)$ possible edges with probability $p = e/ (n^2-n)$, assigning them a random weight between $1$ and $100$,
and adding randomly connected intake and excretion nodes. 
We then use the algorithms from \cite{merrill2019stability} to determine the existence of an equilibrium and its metabolite concentrations. Determining the existence of an equilibrium relies on 
a topological argument and a Dijkstra algorithm provided in \cite{merrill2019stability}, while finding the concentrations relies on inverting a grounded Laplacian. This approach is described in Appendix \ref{appendix:maths}.
This allows us to create the large sets of problems and solutions required to train AI models. For each problem, we create four datasets with over $40$ million examples of graphs of varying number of nodes: 8-32 (S), 32-64 (M), 64-128 (L) and 128-256 (XL).

As the number of nodes increases, the proportion of graphs with an equilibrium goes down. However, our models require balanced datasets ($50\%$ with an equilibrium, $50\%$ without) for the qualitative problem, and graphs with an equilibrium for the quantitative problem (there is no equilibrium to predict otherwise). To remedy this, we use a redemption technique which modify graphs without an equilibrium until they admit one (see Section~\ref{sec:discussion} and Supplementary~ \ref{appendix:datasets}).

\begin{figure}[h!]
    \centering
    \includegraphics[width = 1.0\textwidth]{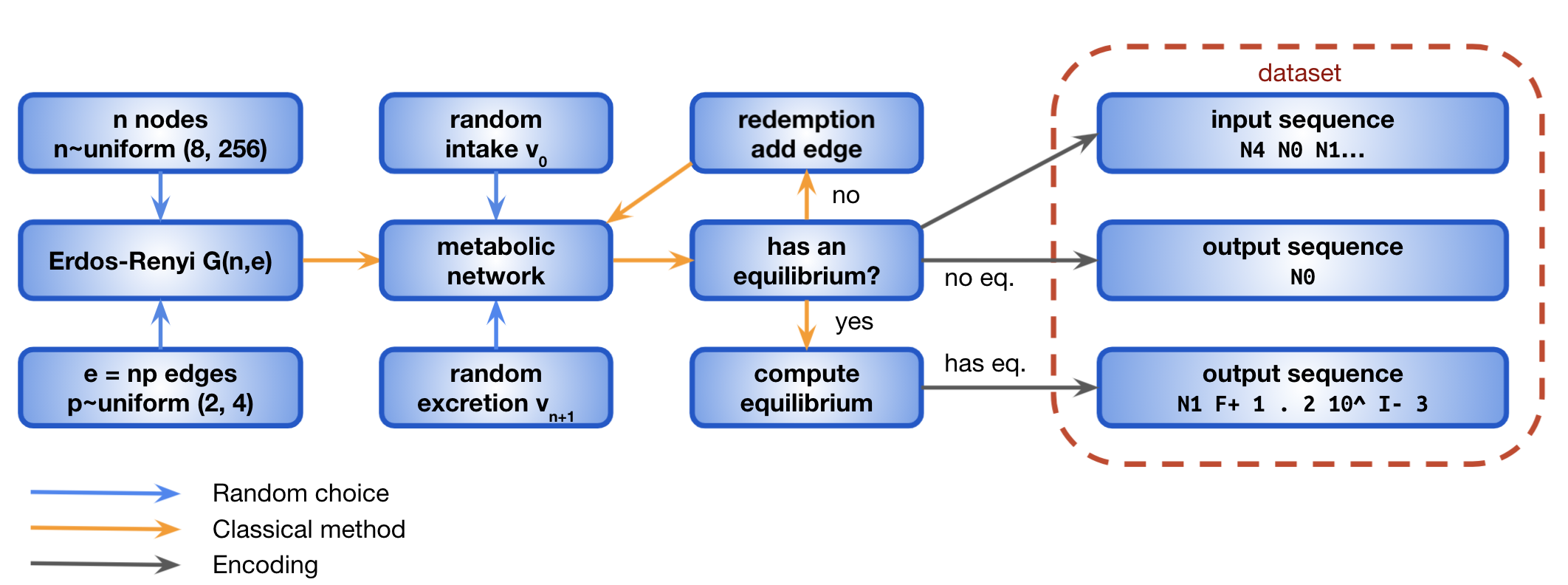}
    \caption{Generation procedure for synthetic networks\label{fig:generation}}
    
\end{figure}

For our models, we use Transformers \cite{transformer17},
a deep learning architecture originally designed for natural language processing and machine translation \cite{lample2017unsupervised}, but that was recently applied to such diverse tasks as symbolic mathematics \cite{LampleCharton}, computation \cite{charton2021learning} and computer vision \cite{carion2020endtoend} (see section~\ref{sec:methods}). Transformers are sequence to sequence models: they process sequences of input tokens to produce sequences of output tokens. In our datasets, the input are graphs, represented as sequences of pairs of symbolic tokens corresponding to edges, with an additional token for the edge weight in the quantitative case (see Figure \ref{fig:encode}). Output sequences are limited to one token in the qualitative case: $0$ for ``has no equilibrium" and $1$ for ``has an equilibrium". In the quantitative case, they are a series of numbers (the concentrations in every node) written in scientific notation and represented as sequences of digits, decimal point, sign and exponents. The generation process is summarized in Figure~\ref{fig:generation}.

\subsection{Predicting equilibria}
In qualitative experiments, the model is trained to predict a binary output: whether the graph has an equilibrium or not. Trained model are tested on held-out samples of graphs not encountered during training. For all datasets (i.e. all graph sizes), the existence of an equilibrium is predicted correctly in more than $98.9\%$ of cases. Shallow models, with only one layer in the encoder and decoder, are sufficient for qualitative tasks. Our best performing models have $256$ dimensions and $32$ attention heads, but much smaller models ($64$ dimensions and $8$ heads) achieve over $94\%$ accuracy as well. Table~\ref{tab:quali_results1} summarizes our results. 
\begin{table}[h]
    \small
    \centering
    \begin{tabular}{lcc}
        \toprule
        & Best (256/32) &  Low dimension (64/8)\\
        \midrule
        Small & 98.9 &  98.2 \\
        Medium & 99.6 & 99.4 \\
        Large & 99.5 & 95.9 \\
        Extra-large & 99.3 & 94.1 \\
        \bottomrule
    \end{tabular}
    \caption{\small \textbf{Qualitative experiments, (dimension/heads)}All models have one-layer encoder and one-layer decoder.}
    \label{tab:quali_results1}
\end{table}

In quantitative experiments, the model predicts a vector of floating point numbers, corresponding to the concentrations in each node at equilibrium. Accuracy is measured as the proportion of predictions that are both syntactically correct
(i.e. predict one concentration per node)
and fall within a certain tolerance of the correct mathematical solution. 
This tolerance is defined as a relative error in $l_{1}$ norm and is typically chosen to be $10\%$, $5\%$, $2\%$ or $1\%$ depending on the experiment. 
On the small and medium graph data sets, transformers with $8$ encoding layers, $2$ or $4$ decoding layers, $512$ dimensions and $8$ attention heads can predict equilibrium up to a tolerance of $10\%$ in more than $99.8\%$ of the test cases. At stricter tolerance levels ($5, 2$ or $1\%$), accuracies of $99, 96$ and $82\%$ are achieved, but learning curves (see Figure~\ref{fig:training_quanti} in Supplementary~\ref{appendix:detailed_results}) suggest that longer training time would bring all accuracies close to $100\%$. Over larger graphs ($64$ to $128$ nodes), transformers can be trained to $98.8\%$ accuracy as well, and $96.6\%$ at $5\%$ tolerance, but the very long sequences ($900$ tokens on average for input and output) make training very slow, and tighter tolerance levels would need a lot of time to be learnt. On graphs with more than $128$ nodes, models become very difficult to train (for more results and experiments with unweighted graphs, see Supplementary~\ref{appendix:detailed_results}). 

\begin{table}[h]
    \small
    \centering
    \begin{tabular}{lcccc}
        \toprule
         & $10\%$  &  $5\%$ & $2\%$ & $1\%$ \\
        \midrule
        Small (8/2/512/8) & 99.8 & 99.4  & 96.8 & 87.9 \\
        Medium (8/4/512/8)  & 99.9 & 99.2 & 96.2 & 82.8 \\
        Large (8/4/512/8) & 98.8 & 96.6 & 81.5 & 50.0  \\
        \bottomrule
    \end{tabular}
    \caption{\small \textbf{Quantitative experiments, accuracies for different tolerance} level. Model sizes are (encoder layers/decoder layers/dimensions/attention heads).}
    \label{tab:quanti_results1}
\end{table}

\subsection{Out-of-distribution generalization and results on real metabolic networks}
\label{subsec:outandreal}
All the results so far were calculated on graphs that the model had not seen during training, but that were generated with the same methods as the training data. They have the same number of nodes, the same ratio of edges to nodes (between $2$ and $4$) and the same Erd\H{o}s-R\'enyi distribution. We believe these choices of parameters make sense for metabolic graphs (see section~\ref{sec:discussion}). However, it is important to assess how much our predictions depend on dataset generation procedures, and what accuracy can be achieved on graphs from different models, with different properties. 
This is known as the out-of-distribution generalization problem. To address it, we tested trained models for the qualitative and quantitative problems on sets of graphs with different number of nodes, differents proportion of edges ($1-2$, or $4-5$ vs $2-4$), and generated from different graph models (small world and scale free instead of Erd\H{o}s-R\'enyi \cite{watts1998collective,bollobas2003directed}).

Over graphs with different edge to node ratios (denser or more sparse), or generated using different models, our trained models achieve high accuracy (over $90\%$) on all tasks. Sometimes, the results are even comparable to those achieved over the training distribution. 
Testing on graphs of different size, qualitative models can predict the existence of equilibrium on larger graphs to some accuracy ($90$ and $80\%$ for M-trained models on L and XL graphs), but fail to generalize to graphs with fewer nodes. Quantitative models do not generalize at all to graphs of different sizes, probably because their output size strongly correlates with graph length (for a graph with $n$ nodes, the output has about $10n$ tokens). Results on medium sized models are summarized in table~\ref{tab:generalization1}, detailed results can be found in Supplementary~\ref{appendix:generalization}. 

These are important findings: even if real-world metabolic graphs are slightly different from those we use for training, our models will nevertheless correctly predict their qualitative and quantitative properties, so long the training set includes graphs with the same number of nodes. 
\begin{table}[h]
    \small
    \centering
    \begin{tabular}{lcc}
        \toprule
         & Qualitative  &  Quantitative  \\
         & chance level $50\%$ & chance level $0\%$ \\
        \midrule
        \textbf{Proportion of edges} \\
        1 - 2 & 100 & 97.7 \\
        4 - 5  & 96.0 & 92.8 \\
        \midrule
        \textbf{Different generators} \\
        Small world & 99.2 & 96.5 \\
        Scale free  & 99.4 & 97.0 \\
        \bottomrule
    \end{tabular}
    \caption{\small \textbf{Out of distribution generalization, for models trained on graphs with 32-64 nodes, 2-4 edges per node, and Erd\H{o}s-R\'enyi distribution.}}
    \label{tab:generalization1}
\end{table}

Finally, we evaluated our trained models on experimental metabolic
networks from the KEGG database \cite{10.1093/nar/28.1.27,10.1093/nar/gkaa970}. 
Biological pathways were extracted by selecting all pathways having an equilibrium and involving a chemical reaction, identifying intakes as nodes with all edges pointing outward edges and excretion as nodes with all edges pointing inward, and connecting them to virtual intake and excretion nodes. This provided a test set of 29 networks with $4$ to $40$ nodes, that includes such metabolisms as lysine degradation, vitamin A-1 (retinol), caffeine, nitrogen, taurine and lipoic acid, and the three networks presented in Figure~\ref{fig:graph_examples} (complete list in Supplementary~\ref{annex:kegg_networks}). We trained a qualitative and a quantitative model over random graphs from $4$ to $40$ nodes. After training, the qualitative model achieved $99.7\%$ accuracy over a held-out test set of random graphs, and the quantitative model $99.8, 98.5, 90.8$ and $71.2\%$ accuracy at tolerance $10, 5, 2$ and $1\%$.

When tested on our $29$ biological networks, the qualitative model correctly predicted the existence of an equilibrium in all instances, and the quantitative model predicted the concentrations of $28$ equilibria to $10\%$ tolerance, $27$ to $5\%$, $26$ to $2\%$ and $24$ to $1\%$. These results confirm that models trained on synthetic datasets can predict real biological networks with the same accuracy as generated test sets, provided they are trained on graphs with the same number of nodes as those that will be tested. More specifically, for the three networks from Figure~\ref{fig:graph_examples}, our qualitative model correctly predicted the existence of all three equilibria. The qualitative model predicts the equilibrium for the Reverse Cholesterol Transport and the PBPK medication diffusion model with less than $0.05\%$ error, the theoretical best we can achieve, since it corresponds to the rounding error of our sequence representation. The equilibrium of the cyanobacteria tricarboxilic acid cycle was predicted with $1.25\%$ error.

\section{Methods}
\label{sec:methods}
\begin{figure*}[h!]
    \centering
    \includegraphics[width = 0.95\textwidth]{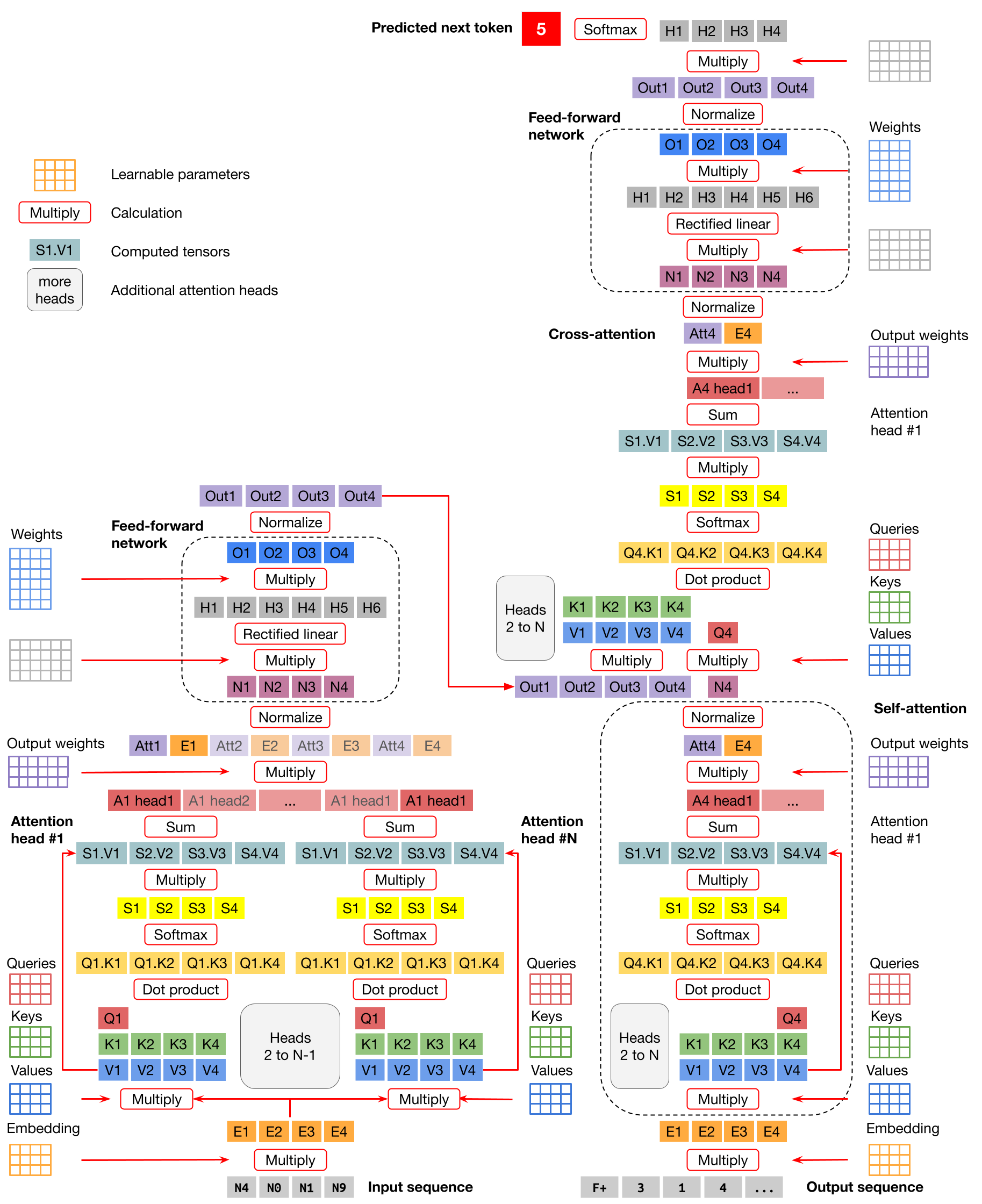}
    \caption{\textbf{The Transformer architecture.} A transformer model with one layer encoder (left) and decoder (right), with the trainable parameters, and the operations performed when the model predicts.}
    \label{fig:architecture}
\end{figure*}

The deep learning architecture leveraged in this research is the Transformer \cite{transformer17}. Transformers process sequences of tokens (or words), which are fed into a multi-layered encoder. In each layer of the encoder, a self-attention mechanism \cite{bahdanau2014neural} takes care of auto-correlations in the sequence, and a feed-forward network is used to process the original sequence and the output of the self-attention layer. Prediction is handled by a multi-layered decoder that has the same architecture, but with uses two attention layers: a self-attention that handles the part of the output sequence already decoded, and a cross-attention that links to the output of encoder. At the top of the decoder, a linear layer and a softmax output the most likely next token, given the input and previously decoded output. Figure \ref{fig:architecture} presents the architecture, the computation process and the trainable parameters of the model.

\begin{figure*}[h!]
    \centering
    \includegraphics[width= 0.95\textwidth]{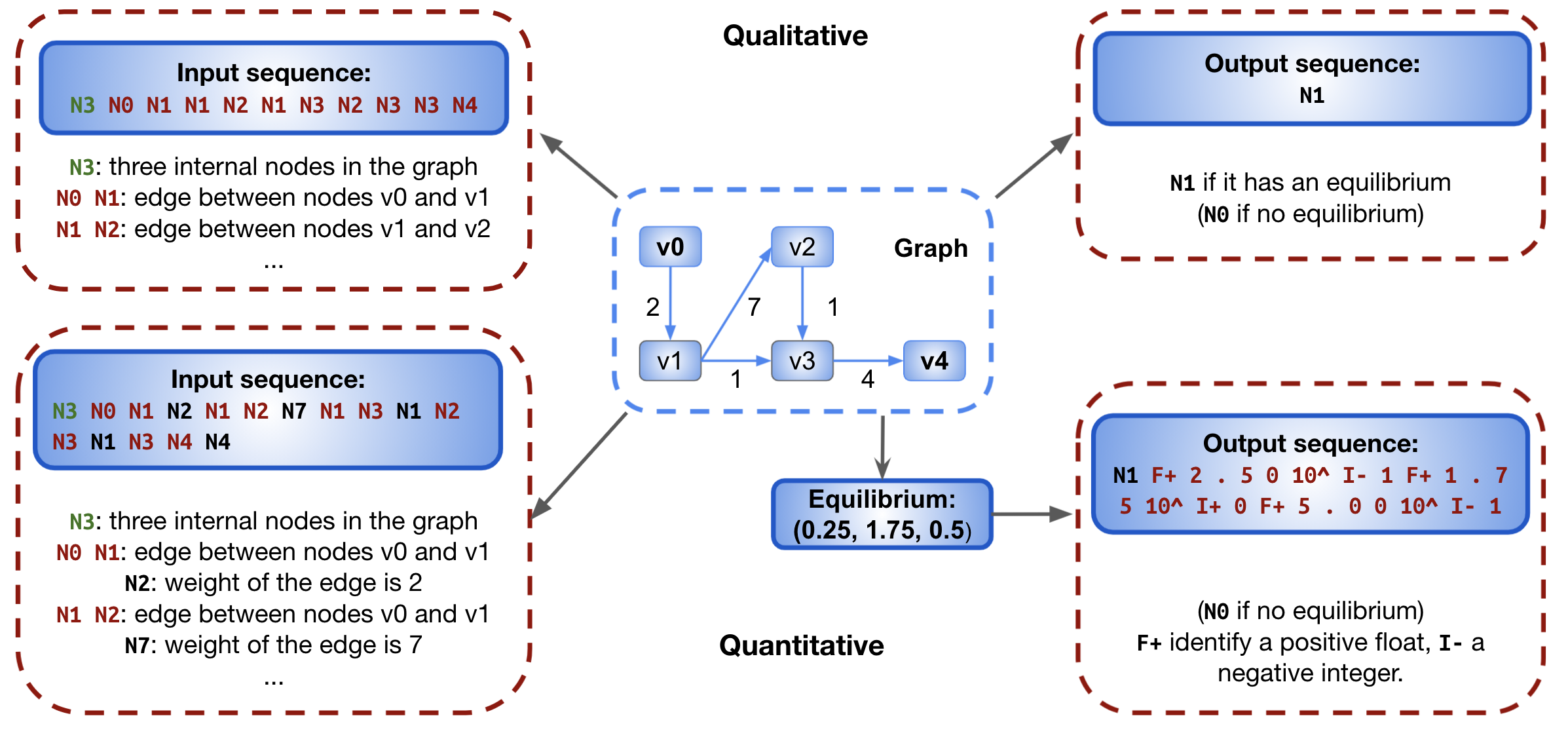}
    \caption{Example of the encoding procedure for the input and output on a 3 nodes graph}
    \label{fig:encode}
\end{figure*}

To encode a directed graph with $p$ nodes as a sequence, we use symbolic tokens $N_1$ to $N_p$ and represent the graph as its number of nodes ($N_p$), and a sequence of pairs representing each edge (e.g. $N_2, N_5$ if nodes $2$ and $5$ are connected). For weighted graphs, we add a third token, from $N_1$ to $N_{100}$, for the weight. The description of our graphs is therefore purely symbolic. When predicting the existence of an equilibrium, the output of our problem is binary, and represented by a single token, $N_0$ if there is no equilibrium or $N_1$ if there is an equilibrium. When predicting values at equilibrium, the output of a vector of $p$ real numbers, represented as a sequence of digits (e.g. '+', '1', '.', '2', '1', '10\^', '-', '1', 'separator' ...). All numbers are written in floating point notation and rounded to three significant digits. This encoding is represented in Figure \ref{fig:encode}

During training, the loss function we try to minimize is the cross-entropy of the solution from the training dataset, with respect to the distribution of tokens generated by the model. This is the standard technique in natural language processing. It amounts to minimizing the Kullback-Leibler divergence between the distribution of output tokens predicted by the model and that of the dataset (i.e. making the model output tokens with the same distribution as the training set). It is important to note that the meaning of tokens, here their mathematical significance, has no impact on the training loss. During training, the model uses no information on the mathematical nature of the problem, or even the fact that is it decoding numbers. At test time, on the other hand, the output sequences are decoded and interpreted as real numbers, and the model prediction is considered correct if it is a valid representation of a real vector and if the $l_1$ norm of its difference with the solution is less than a percentage of the $l_1$ norm of the solution. We consider tolerances of $10, 5, 2$ and $1\%$.

The code and the synthetic data generated will be made openly available as well as the trained models used in this paper.

\section{Discussion}
\label{sec:discussion}
The very high accuracy achieved by our trained models suggests that they can serve as reliable predictors of graph equilibrium. Our experiments with out-of-distribution generalization show that they predictions remain accurate even when tested on graphs with different structures or edge distributions than those used to train the models. Our first experiments on a limited set of biological graphs 
confirm these findings.

Our approach can scale to larger graphs. Trained models predict the equilibrium of small and large networks with the same accuracy, provided that they are trained on graphs with the same number of nodes as those they predict. As we have seen, this is necessary for our models to generalize, but since we generate our training data and can create graphs of any size, it is a very weak constraint in practice. Yet, it sets a limit to our approach: the largest equilibrium we can predict is the largest graph we can train on, and this is constrained by the transformer architecture, because the attention mechanism is quadratic in the length of the sequence (in speed and memory usage). In our experiments, using $4$ V100 GPU with $32$ gigabytes of memory each, training became very slow and unstable once inputs or outputs exceeded $2000$ tokens.
In the qualitative problems, lengths of input is twice the number of edges, or eight times the number of nodes, and output length is always one. This sets the graph limit size around $250$ nodes. For the quantitative problem, input length is three times the number of 
edges, or twelve times the number of nodes, and output (with three significant digits) ten times the number of nodes, which sets the limit around $170$ nodes. This is however 
above the size of many relevant
biological networks, as for instance those selected from the KEGG database.
Since memory requirements increase as the square of sequence lengths, adding more computational power will only help with graphs slightly above this limit. For larger graphs, we could use compact representations of input and output, or introduce recent transformer architectures, like sparse attention models \cite{child2019generating} and other techniques that remove the quadratic dependence on sequence length. This is an active research topic in machine learning \cite{zaheer2021big, wang2020linformer, vyas2020fast}, making it likely that this limitation will be lifted in the near future.

All our models are trained from randomly generated graphs with $2$ to $4$ edges per node and equiprobable edges. This is known as the Erd\H{o}s-R\'enyi model, a standard and minimal assumption \cite{gilbert1959random,erdHos1960evolution}.
Our experiments with out-of-distribution generalization suggest that training on graphs with an Erd\H{o}s-R\'enyi distribution does not hinder the prediction of graphs from different models (small world or scale free \cite{watts1998collective,bollobas2003directed}) or edge density (less than two or more than four edges per node). 

Graphs with an equilibrium become increasingly rare as their number of nodes grow: the proportion is $15\%$ for $50$ nodes, $3.5\%$ for $100$, $0.9\%$ for $150$ and $0.3\%$ for $200$ nodes. To train our qualitative models, we need $50\%$ of the examples in the dataset to have an equilibrium, and all of them in the quantitative case. As  a result, we need to over-sample graphs with an equilibrium. The typical procedure for this, rejection sampling (generating randomly until we have enough graphs with an equilibrium, and throwing away the supernumerary examples), would be extremely inefficient for large graphs. To create a sample of $40$ millions graphs with $100$ nodes, using rejection sampling, we would need to generate $570$ million graphs in the qualitative case, and $1.1$ billion in the quantitative (these figures would be $2.2$ and $4.4$ billions for graphs with $150$ nodes). Here, we use a redemption method which builds an equilibrium from a graph having none. For graphs with $100$ nodes, it reduces the generation time by a factor of $3$ (and a factor of $7$ for $150$ nodes) at the price of a small departure from the Erd\H{o}s-R\'enyi distribution. We have tested models built on rejection sampled graphs (see Supplementary~\ref{appendix:detailed_results}) and found no difference in accuracy. In fact, the redemption-sampled datasets performed better in the generalization experiments. This might be due to the additional diversity introduced by the redemption procedure, a phenomenon that was mentioned in \cite{LampleCharton}.

Experiments with model architectures reveal important differences between this problem and traditional language tasks. In language processing, there is a strong tendency to resort to very deep architectures \cite{brown2020language}, and, due to the symmetrical structure of translation problems, to have transformers with the same number of layers in the encoder and decoder. In our experiments, we observe that shallow decoders are always sufficient. In fact, decoders with only one layer will usually achieve performances close to the best models (i.e. over $99.5\%$ accuracy). For the quantitative problem, deep encoders are necessary. In the qualitative case, high accuracy can always be achieved with one-layer encoders, and very small models, with $64$ dimensions and $8$ attention heads, are almost as good as the best on graphs with up to $64$ nodes, and achieve accuracy over $94\%$ for the larger cases.
Such performance of shallow and small models, which was already noted in \cite{charton2021learning}, is unheard of in natural language processing. The efficiency of shallow models (or decoders) is an interesting feature, since it contributes to making models smaller, easier to train, even without large clusters or HPC resources, and faster at inference. 

The high performance of transformers trained on generated datasets for solving these two metabolic graph problems suggests that this approach could be applied to related problems, such as ensuring the stability of the equilibria through Lyapunov
functions \cite{AAS20}, extreme pathways
\cite{OTP2010} or Flux Balance Analysis (FBA) as a general tool \cite{palsson2015}. 
There is evidence of the use of such
tools also for dynamic problems
\cite{charton2021learning} thus opening
the door to address dynamic FBA \cite{MAHADEVAN2002}.
More generally, the approach
is also a promising step toward using such efficient AI models for dynamic problems
in wider areas, such as 
drug discovery through Quantitative Systems Pharmacology and PBPK models \cite{QSPH21}, or synthetic biology with flux optimization \cite{SB06,Collins2010}. 

\section{Conclusion}
We have shown that transformers can be trained on randomly generated graphs to predict the qualitative and quantitative properties of biological metabolic networks. Our results scale to graphs with up to 256 nodes, and generalize out of their training distribution to networks with different connectivity factors, or models (e.g. small world models). Compared to models typically used for natural language processing, we rely on small architectures, with different depth for the encoder and decoder. 

Our approach constitutes a proof of concept. We believe it is a first step towards larger use of natural language models, and especially transformers, as an end-to-end solution to solve problems of computational biology.

\section*{Acknowledgments}
The authors would like to thank Yann Ollivier and Guillaume Lample for insightful discussions. 
S.M., N.M. and B.P. would like to acknowledge the support of the Joseph and Loretta Lopez Chair endowment.
Finally the authors would like to thank the French Corps des IPEF.
\nolinenumbers

\newpage
\bibliographystyle{plos2015.bst}
\bibliography{bibliography}

\clearpage

\appendix
\newpage
\section{Classical methods and mathematical representation}
\label{appendix:maths}
The classical computation methods used to solve the qualitative and quantitative problems considered here (see for instance \cite{merrill2019stability}) are based on the equations of the metabolite dynamics (see \cite{gunawardena2012linear,NEWLIFE}).
Let us denote 
$v_{0}$ 
the level of the virtual intake node, $x$ the concentrations of metabolites and $f$ the exchange flux, the general metabolic dynamics is given by
\begin{equation}
    \frac{dx}{dt} = F(x,f,v_{0}).
\end{equation}
In general $f$ has itself a dynamic that depends on $x$ and $f$. However, it is usually assumed that the exchange flux $f$ have a very slow variation compared to the metabolite dynamics and can therefore be considered as constants \cite{gunawardena2012linear,klinke2012timescale}. In this case, an equilibrium is a state of the metabolites $x_{e}(f,v_{0})$ depending on the exchange flux $f$ and the intake $v_{0}$ such that
\begin{equation}
F(x_{e}(f),f,v_{0})=0.
\end{equation}
When the dynamics of the metabolites is linear in $(v_{0},x)$, then the dynamics becomes (see for instance \cite{gunawardena2012linear})
\begin{equation}\label{eq:dyn1}
    \frac{dx}{dt} = J_{1}(f)x+\phi v_{0},
\end{equation}
where $\phi$ is a column vector corresponding to the flux of the intakes and $J_{1}(f)$ the jacobian matrix of $F$ with respect to $x$. 
and therefore an equilibrium $x_{e}(f,v_{0})$ is a solution to 
\begin{equation}
    J_{1}(f)x_{e}(f,v_{0}) = -\phi v_{0}.
\end{equation}
In this case if a unique equilibrium exists then $J_{1}(f)$ is invertible and 
\begin{equation}
    x_{e}(f,v_{0}) = -J_{1}(f)^{-1}\phi v_{0}.
\end{equation}
$L = -J_{1}(f)^{T}$ is called a grounded Laplacian and can be directly obtained 
by taking the adjacency matrix $A$ of the subgraph where the intake node $v_{0}$ is removed, then defining $D$ the diagonal matrix whose entries are the row sums of $A$, computing $(A-D)$. Then removing the last rows and column to $(A-D)$ gives $J_{1}(f)^{T}$. This is the procedure mentioned in Section \ref{sec:results} (see also \cite{merrill2019stability}).
\subsection{Erd\H{o}s-R\'enyi, small world and scale free graph models}
The networks we use as training data are generated using the $G_{n,p}$ algorithm first introduced in \cite{gilbert1959random}. For each graph, the number of nodes and edges are selected randomly from a discrete uniform distribution, and each of the $n(n-1)$ possible edges is selected with probability $p = e/(n^2-n)$. This generates a distribution where all graphs with $n$ nodes and $e$ edges are equiprobable ($n$ and $e$ being themselves uniformly distributed). An alternative version known as $G_{n,m}$ introduced in \cite{erdHos1960evolution} select uniformly at random a graph among all possible graphs with $n$ nodes and $m$ edges generates the same distribution for a given number of nodes and edges. Generating all graph with uniform distribution is a very neutral assumption, which makes Erd\H{o}s-R\'enyi a good candidate for our training set generation.

Other graph models exist that generate networks with specific properties, that would be very unlikely in graphs generated by the Erd\H{o}s-R\'enyi algorithm. The small world model generates graphs with short average path lengths between nodes in the graph, and a large clustering coefficient. These correspond to many real-world networks, and can be created using the Watts–Strogatz model \cite{watts1998collective}. Since the original Watts–Strogatz model is an undirected graph model, we modified it to generate directed graphs. Scale free graphs, another important class, have very large clusters, which would have a very small probability of appearing in Erd\H{o}s-R\'enyi models. Such graphs typically have a number of nodes with $k$ connections behaving like $ C k^{-\gamma}$ for large $k$ where $C$ and $\gamma$ are some positive constants, while for Erd\H{o}s-R\'enyi this number would decay much faster. They can be generated with a generation algorithm for scale free directed graph adapted from \cite{bollobas2003directed}.
Scale free and small world graphs were generated to test out-of-distribution generalization (see Section \ref{subsec:outandreal} and Supplementary \ref{appendix:generalization}).

\section{Datasets \label{appendix:datasets}}
We generate data for two problems: predicting whether an equilibrium exists (the qualitative problem) and predicting the flows in each node at equilibrium (the quantitative problem). In the latter, graph edges have associated weights to account for the properties of the reactions they represent, but we also consider an unweighted case, where all weights are set to one (for lack of better information). For each three cases, we created four sets of graphs, with different number of nodes: 8-32 (S), 32-64 (M), 64-128 (L) and 128-256 (XL). Since XL weighted graphs were too large to train efficiently, only eleven datasets were created. 

Samples for the qualitative problems are balanced, so that $50\%$ of the graphs have an equilibrium. For the quantitative problems, only graphs with an equilibrium are considered. Because the proportion of graphs with an equilibrium goes down as the size of the graphs increase, naive rejection sampling would create data sets heavily biased towards small graphs. Rejection sampling from the conditional distribution (first selecting the number of nodes, and then deciding whether we want a graph with equilibrium, with probability $0.5$ in the qualitative case, and $1$ for the quantitative) will eliminate this bias, but will be extremely slow for large graphs. Instead, we use a redemption procedure that creates an equilibrium from a graph that lacks one by selecting a node at random, among those not connected to the output node, connecting it to the output node, and iterating until an equilibrium is found. To build a dataset, we generate random graphs, and ``redeem'' graphs without equilibrium according to a certain probability ($1.0$ in the quantitative problems, from $0.05$ to $0.49$, depending on graph size in the qualitative case). This allows data to be generated with acceptable speed. To make sure that our redemption procedure introduces no bias in the data, we generated an additional ``filtered'' dataset of M-sized graphs (32 to 64 nodes), using conditional rejection sampling. 

In the qualitative and unweighted datasets, input sequences consist of a list of pairs of symbolic tokens, describing the edges of the graph (e.g. the pair $(N1, N3)$ designate an edge from node $1$ to node $3$). In the weighted cases, we need a third value to indicate the weight, between $1$ and $100$. To limit sequence length, we encode it using the same symbolic tokens as the nodes ($N1$ to $N100$). To make sure that this has no impact, we created an M-sized dataset where weights are encoded as sequences of digits instead (from $['+', '1']$ to $['+', '1', '0', '0']$).

In quantitative problems, output sequences represent vectors of floating point numbers. Since the tightest tolerance we consider is $1\%$, we round them to three significant digits in the mantissa. To assess the impact of rounding in the output, we created an additional M-sized dataset with solutions rounded to four significant digits. Table~\ref{tab:data_stats} summarizes the 18 datasets that were generated.

\begin{table}
    \small
    \centering
    \begin{tabular}{lcccccc}
        \toprule
        & Sample & Nr & Average & Average & Max & \\
        & size & nodes &input& output & length & Redeem\\
        \midrule
        \textbf{Qualitative}\\
        Small & 38.9M & 8 - 32 & 124 & 1 & 256 & 0.05\\
        Medium & 49.0M & 32 - 64 & 295 & 1 & 512 & 0.4 \\
        Medium Reject sampling & 44.4M & 32 - 64 & 306 & 1 & 512 & 0.4 \\
        Large & 41.4M & 64 - 128 & 588 & 1 & 1024 & 0.48 \\
        eXtra Large & 31.5M & 128 - 256 & 1173 & 1 & 2048 & 0.49 \\
        \midrule
        \textbf{Quantitative}\\
        Small & 48.4M & 8 - 32 & 128 & 181 & 608 & 1\\
        Medium & 33.7M & 32 - 64 & 302 & 433 & 1216 & 1 \\
        Medium Reject sampling & 33.8M & 32 - 64 &  330  & 407 & 1216 & 1 \\
        Medium 4 digits & 33.3M & 32 - 64 & 302 & 481 & 1300 & 1 \\
        Large & 32.9M & 64 - 128 & 599 & 865 & 2432 & 1 \\
        eXtra Large & 31.5M & 128 - 256 & 1195 & 1729 & 4864 & 1 \\
        \midrule
        \textbf{Weighted quantitative} \\
        Small & 49.2M & 8 - 32 & 192 & 181 & 717 & 1\\
        Medium & 59.6M & 32 - 64 & 452 & 433 & 1436 & 1 \\
        Medium Reject sampling & 64.0M & 32 - 64 &  495  & 407 & 1450 & 1 \\
        Medium 4 digits & 76.3M & 32 - 64 & 452 & 481 & 1539 & 1 \\
        Medium Num weights & 69.6M & 32 - 64 & 739 & 433 & 2176 & 1 \\
        Large & 39.0M & 64 - 128 & 898 & 865 & 2870 & 1 \\
        \bottomrule
    \end{tabular}
    \caption{\small \textbf{Training set sizes, number of nodes, average input and output length (in tokens), maximum length (input plus output) and redeem probability for different datasets.} Reject sampl: sample built by conditional rejection sampling (vs redeeming). 4 digits: output saved with 4 significant digit (vs 3). Num weights: edge weights as digit sequences (vs symbolic tokens).}
    \label{tab:data_stats}
\end{table}

\section{Detailed results \label{appendix:detailed_results}}
\subsection{Qualitative problem}
In these experiments, we predict a binary output: whether the graph has an equilibrium. We trained transformers with $1$ to $4$ layers, $8$ to $32$ attention heads, and embeddings with $64$ to $256$ dimensions. For all datasets, our models achieve accuracies around or over $99\%$, on held-out test data. This can be done with shallow transformers, with only one layer in the encoder and decoder. Deep models bring no increase in accuracy. Our best models use $32$ attention heads and $256$ dimensions, but smaller models, with only $64$ dimensions and $8$ attention heads, are almost as good on small and medium sized graphs (up to $64$ nodes), and achieve accuracies over $94\%$ even in the larger cases. That such shallow and small models correctly predict comes as a surprise: they could not be used even for simple language processing tasks. 
Models trained on medium sized graphs generated with conditional rejection sampling show no significant difference in accuracy, suggesting that our balance-by-redemption method has no impact on our results. Table~\ref{tab:quali_results} summarizes the results. 
\begin{table}[h!]
    \small
    \centering
    \begin{tabular}{lcc}
        \toprule
        & Best (256/32) &  Low dimension (64/8)\\
        \midrule
        Small & 98.92 &  98.18 \\
        Medium & 99.55 & 99.39 \\
        Large & 99.51 & 95.89 \\
        Extra-large & 99.26 & 94.10 \\
        \midrule
        Medium reject* & 99.66 & 99.63 \\
        \bottomrule
    \end{tabular}
    \caption{\small \textbf{Qualitative experiments, (dimension/heads)}All models have one layer encoder, and one layer decoder. Medium reject best model is 256/16.}
    \label{tab:quali_results}
\end{table}

\subsection{Quantitative problems}
\begin{figure}[h!]
{\centering \includegraphics[width=0.7\linewidth]{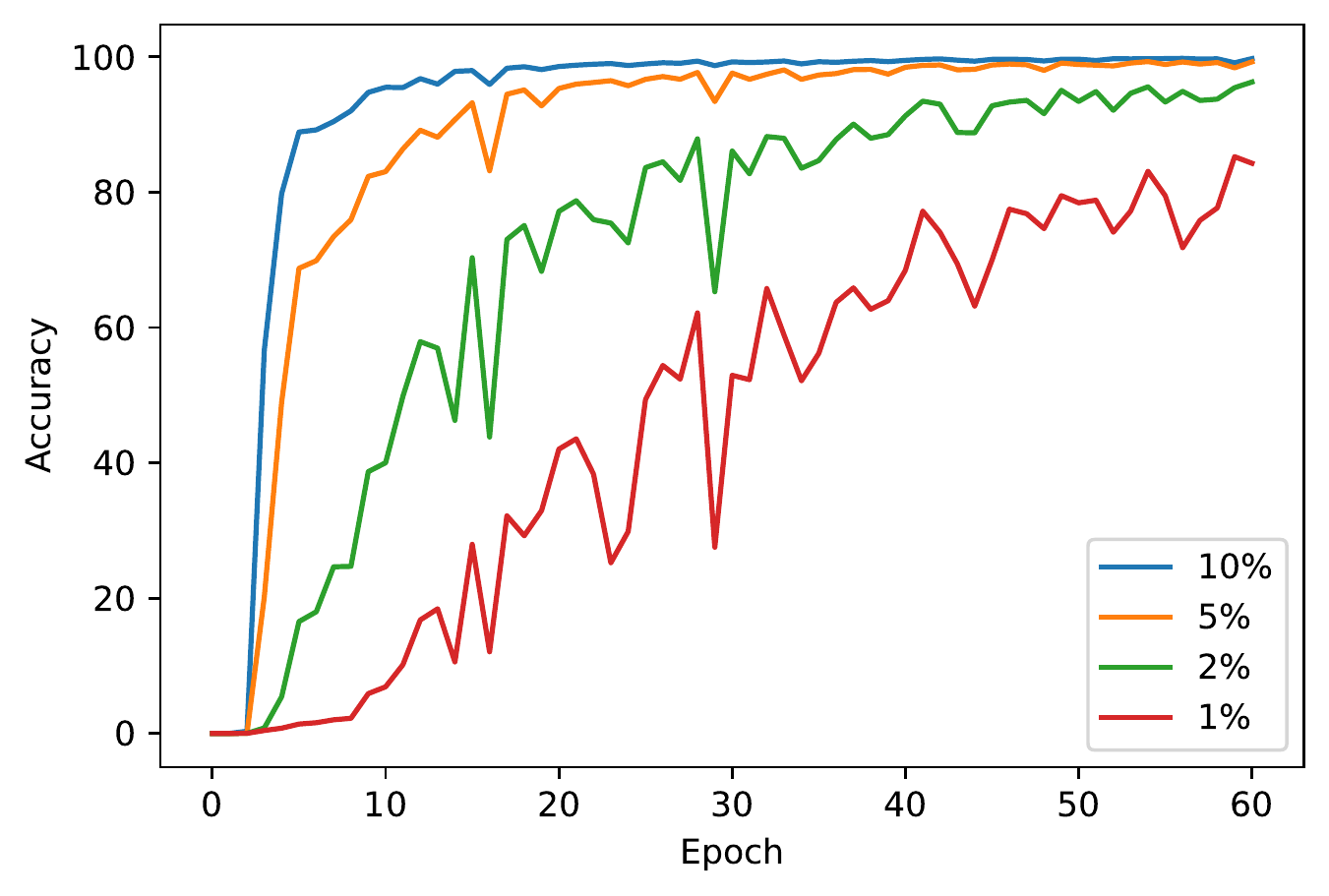}\par}
\caption{\small \textbf{Learning curves for different tolerance levels.} 8/2/512/8 transformer trained over small graphs.}
\label{fig:training_quanti}
\end{figure}

In these experiments, we predict metabolite concentrations at equilibrium, a vector of floating point numbers represented as sequences of digits and symbols. The input sequence is the same as in the qualitative problem, but output length is proportional to the number of nodes. At test time, accuracy is defined as the proportion of valid predictions (i.e. sequences that can be decoded as a vector of floats) that fall within a certain tolerance of the correct mathematical solution (i.e. such that ratio the $l_1$ norm of the difference between the prediction and the solution to the $l_1$ norm of the solution is below some fixed tolerance level). We use a tolerance of $10\%$ as our main metric, but also provide results for $5, 2$ and $1\%$. We consider two cases: weighted graphs, which have flow values associated to each edge, and unweighted graphs, where all edges have the same weights (conventionally $1$). This will happen in some transport graphs, or as a neutral hypothesis when the actual weights of edges are unknown.

in the unweighted case, on the small, medium and large data sets, transformers with $8$ encoding layers, $2$ decoding layers, $512$ dimensions and $8$ attention heads can predict equilibrium with less than $10\%$ error on more than $99.5\%$ of the test cases. At stricter tolerance levels ($5, 2$ or $1\%$), we achieve $99, 96$ and $86\%$ accuracy respectively, but as Figure~/\ref{fig:training_quanti}) suggests, higher accuracy and stricter tolerances can be reached with more training. For the largest graphs ($128$ to $256$ nodes), transformers can be trained to $85\%$ accuracy, but the very long sequences ($2900$ tokens on average, with a maximum at $4864$) make memory requirements very high, and the training very slow and prone to failure. Using conditional rejection rather than our redemption procedure, or providing the model with outputs with more significant digits for training purpose, have no noticeable impact on accuracy for medium sized graphs. 
Tables~\ref{tab:quanti_results} and \ref{tab:quanti_ablation_depth} summarize our results.
\begin{table}[h]
    \small
    \centering
    \begin{tabular}{lcccc}
        \toprule
         & $10\%$  &  $5\%$ & $2\%$ & $1\%$ \\
        \midrule
        Small (8/1/512/8) & 99.82 &  99.40 & 96.65 & 86.20\\
        Medium (8/2/512/8)  & 99.67 & 99.23 & 97.17 & 90.48 \\
        Large (8/2/512/8) & 99.85 & 99.50 & 97.15 & 88.75 \\
        Extra-large (8/8/512/8) & 84.70 & 62.80 & 25.50 & 5.80 \\
        \midrule
        Medium reject (8/2/512/8) & 99.78 & 99.60 & 98.06 & 90.57 \\
        Medium 4 digits (8/2/512/8) & 99.71 & 99.21 & 96.73 & 88.29 \\
        \bottomrule
    \end{tabular}
    \caption{\small \textbf{Unweighted quantitative experiments, accuracies for different tolerance}}
    \label{tab:quanti_results}
\end{table}

The best accuracies for small, medium and large graphs are reached with deep encoders (8 layers) and shallow decoders (1 or 2 layers). Studying the impact of the number of layers in encoder and decoder (table~\ref{tab:quanti_ablation_depth}), we observe that whereas having a deeper encoder usually results in higher accuracy, for a given encoder depth, one or two-layer decoders usually perform as well as deeper architectures, and are easier to train. Increasing the number of attention heads does not improve accuracy. 

\begin{table}[h]
    \small
    \centering
    \begin{tabular}{lcccc}
        \toprule
         & $1$-layer dec.  &  $2$-layer dec. & $4$-layer dec. & $8$-layer dec. \\
        \midrule
        \textbf{Small graphs} \\
        1-layer encoder & 25.7 & 66.7 & 90.3 & 96.6 \\
        2-layer encoder  & 87.4 & 94.1 & 99.0 & 99.0 \\
        4-layer encoder & 98.9 & 98.2 & 97.9 & 98.9 \\
        8-layer encoder  & 99.8 & 99.7 & 99.7 & 99.3 \\
        \midrule
        \textbf{Medium graphs} \\
        1-layer encoder & 3.5 & 13.4 & 33.2 & 41.2 \\
        2-layer encoder  & 43.0 & 45.4 & 69.9 & 40.6 \\
        4-layer encoder & 92.9 & 94.6 & 91.3 & 91.1 \\
        8-layer encoder  & 99.8 & 99.7 & 99.3 & 99.5 \\
        \bottomrule
    \end{tabular}
    \caption{\small \textbf{Unweighted quantitative experiments, accuracies for different depths}}
    \label{tab:quanti_ablation_depth}
\end{table}

We represent weighted graphs by adding a (symbolic) token, $N_1$ to $N_{100}$ to each edge. This makes the problem more difficult, and larger models, with deeper decoders (4-layers) and needed. However, the same levels of accuracy are reached after training, and  our best model predict the new equilibrium in almost all cases (Table~\ref{tab:quanti_weighted_results}). Even though the models considered here are larger than those used for the qualitative and unweighted problems, they are much smaller than those used in natural language. The largest architectures used here have less than $50$ million trainable parameters, whereas a recent large language model like GPT-3 has 175 billions. As previously, using the slower rejection sampling approach to produce the dataset, or providing the model with more accurate solutions during training, have no impact on accuracy. Instead of encoding graph weights using symbolic tokens, we also tried to format the as numbers (i.e. strings of digits). This had no impact on accuracy. 

\begin{table}[h]
    \small
    \centering
    \begin{tabular}{lcccc}
        \toprule
         & $10\%$  &  $5\%$ & $2\%$ & $1\%$ \\
        \midrule
        Small (8/2/512/8) & 99.8 & 99.4  & 96.8 & 87.9 \\
        Medium (8/4/512/8)  & 99.9 & 99.2 & 96.2 & 82.8 \\
        Large (8/4/512/8) & 98.8 & 96.6 & 81.5 & 50.0  \\
        \midrule
        Medium reject (8/4/512/8) & 99.9 & 99.6 & 97.9  & 86.0 \\
        Medium 4 digits (8/4/512/8) & 99.7 & 99.2 & 97.4 & 88.6 \\
        Medium num. weights (8/4/512/8) & 99.3 & 96.9 & 85.1 & 59.0   \\
        
        \bottomrule
    \end{tabular}
    \caption{\small \textbf{Weighted quantitative experiments, accuracies for different tolerance}}
    \label{tab:quanti_weighted_results}
\end{table}

\section{Out-of-domain generalization\label{appendix:generalization}}

Using held-out datasets at test time guarantees the our models can generalize to examples they have not encountered during training, but these test exemples are generated with the same procedure as the training data, and belong to the same distribution. In this series of experiments, we consider a stronger form of generalization: the capability of our trained models to correctly predict examples with different characteristics than those encountered during training. We consider three deviations from the training data: larger or smaller networks (i.e. graphs with more or less nodes), different edge density (a different average number of edges per node), and different connectivity models in our graphs (small world or scale free, instead of Erd\H{o}s-R\'enyi). For all 18 datasets, we test the two best models after training on graphs generated with different parameters: sizes from 8-32 to 256-300, number of edges per node 1-2 or 4-5, and small world and scale free graph models.

Qualitative models (Table~\ref{tab:ood_quali_results}) have no difficulty correctly predicting graphs with different edges per node, or structures. The can predict the existence of equilibrium in larger graphs, but fail for smaller graphs and very large ones for small models. 

\begin{table}[h!]
    \small
    \centering
    \begin{tabular}{lcccccccccc}
        \toprule
         & $S_1$  &  $S_2$ & $M_1$ & $M_2$ & $M_{\text{filt}1}$ & $M_{\text{filt}2}$ & $L_1$ & $L_2$ & $XL_1$ & $XL_2$ \\
        \midrule
        8-32 nodes & 98.9 & 98.9 & 50.6 & 50.6 & 50.6 & 50.6 & 49.7 & 49.7 & 46.9 & 47.3 \\
        24-32 nodes & 98.6 & 98.4 & 56.9 & 56.9 & 56.9 & 56.8 & 52.1 & 52.1 & 52.0 & 52.0 \\
        32-40 nodes  & 93.3 & 92.1 & 99.4 & 99.4 & 99.3& 98.0 & 53.6 & 53.6 & 53.6 & 53.6 \\
        32-64 nodes  & 83.7 & 80.2 & 99.6 & 99.5 & 99.2 & 98.3 & 51.6 & 51.6 & 50.2  & 37.2 \\
        56-64 nodes  & 74.3 & 71.1  & 99.7 & 99.7 & 99.7 & 98.9 & 56.8 & 57.0 & 52.1 & 20.3 \\
        64-72 nodes  & 71.1 & 69.3 & 98.7 & 98.7 & 64.4 & 97.7 & 99.9 & 99.9 & 51.2 & 16.2 \\
        64-128 nodes  & 63.9 & 64.5 & 90.1 & 90.0 & 58.4 & 88.8 & 99.5 & 99.5 & 50.9 & 37.8 \\
        120-128 nodes & 58.5 & 62.8 & 86.8 & 83.7 & 56.0 & 82.6 & 99.1 & 99.0 & 61.0 & 55.2 \\
        128-136 nodes & 57.8 & 61.8  & 84.6 & 82.1 & 56.0 & 80.5 & 96.2 & 96.7 & 99.3 & 98.9\\
        128-256 nodes & 54.6 & 60.1 & 78.8 & 75.1 & 54.8 & 74.9 & 87.4 & 83.2 & 99.3 & 99.0 \\
        256-300 nodes & 52.9 & 59.1 & 76.3 & 70.8 & 53.6 & 71.3 & 80.3 & 69.1 & 98.4 & 98.8 \\
        \midrule
        1-2 edges / node & 96.9 & 98.2 & 100 & 100 & 94.8 & 86.5 & 99.8 & 100 & 99.8 & 99.3 \\
        4-5 edges / node & 98.1 & 91.8 & 89.6 & 96.0 & 89.9 & 92.0 & 93.1 & 92.7 & 87.8 & 86.8 \\
        \midrule
        Small-world & 97.7 & 97.7 & 99.2 & 99.2 & 99.0 & 97.5 & 98.9 & 98.7 & 98.4 & 97.8 \\
        Scale free & 84.7 & 89.7 & 99.3 & 99.4 & 97.3 & 83.0 & 99.9 & 100 & 98.5 & 98.8  \\
        \bottomrule
    \end{tabular}
    \caption{\small \textbf{Out-of-domain generalization for qualitative experiments.} (chance level: $50\%$). All qualitative model predictions tend to generalize to graphs with different structure or number of edges. Models tend to generalize to graphs with more nodes than their training set, but not to smaller graphs. $M_{\text{filt}}$: rejection sampling.}
    \label{tab:ood_quali_results}
\end{table}

Quantitative graphs, both weighted and unweighted, generalize to graphs with different structure or edge to node ratio, but fail to predict equilibrium for graphs with different sizes. This is probably due to the fact that the length of output sequences is correlated to the number of nodes. When predicting larger (or smaller graph), the models has to produce a sequence length that never occurred during training. This is shown in Table \ref{tab:ood_quanti_results} and \ref{tab:ood_weighted_results}. In other cases, generalization seems to happen irrespective of the dataset and model, with one exception: models trained on data form rejection sampling seem to have more difficulty. This suggests that the preemption method, by deviating a little from the Erd\H{o}s-R\'enyi procedure, might improve generalization by training on a slightly more diverse dataset. 

\begin{table}[h]
    \small
    \centering
    \begin{tabular}{lccccccccccc}
        \toprule
         & $S_1$  &  $S_2$ & $M_1$ & $M_2$ & $M_{\text{flt}1}$ & $M_{\text{flt}2}$ & $M_{4,1}$ & $M_{4,2}$ &  $L_1$ & $L_2$ & $XL$ \\
        \midrule
        8-32 nodes & 99.8 & 99.8 & 3.8 & 3.8 & 3.4 & 3.3 & 2.7 & 2.7  & 0 & 0 & 0 \\
        24-32 nodes & 99.7 & 99.6 & 11.6 & 11.6 & 10.2 & 10.1 & 8.8 & 8.7 & 0 & 0 & 0 \\
        32-40 nodes  & 10.6 & 10.6 & 99.9 & 99.6 & 88.9 & 88.5 & 97.5 & 97.4 & 0 & 0 & 0 \\
        32-64 nodes  & 2.7 & 2.7 & 99.8  & 99.7 & 87.9 & 87.1 & 99.2 & 99.0 & 2.9 & 2.9 & 0 \\
        56-64 nodes  & 0 & 0 & 99.7 & 99.7 & 86.8 & 85.4 & 98.3 & 97.5 & 11.3 & 11.3 & 0 \\
        64-72 nodes  & 0 & 0 & 11.4 & 11.4 & 10.0 & 9.9 & 11.4 & 11.3 & 99.8 & 99.2 & 0 \\
        64-128 nodes  & 0 & 0 & 1.5 & 1.5 & 1.3 & 1.3 & 1.5 & 1.5 & 99.7 & 99.3 & 0 \\
        120-128 nodes & 0 & 0 & 0 & 0 & 0 & 0 & 0 & 0 & 99.6 & 99.1 & 8.8 \\
        128-136 nodes & 0 & 0 & 0 & 0 & 0 & 0 & 0 & 0 & 11.1 & 11.1 & 84.1 \\
        128-256 nodes & 0 & 0 & 0 & 0 & 0 & 0 & 0 & 0 & 0.8 & 0.8 & 82.7 \\
        256-300 nodes & 0 & 0 & 0 & 0 & 0 & 0 & 0 & 0 & 0 & 0 & 1.5 \\
        \midrule
        1-2 edges/node & 98.7 & 98.7 & 96.8 & 97.5 & 42.9 & 40.7 & 96.4 & 94.5 & 93.7  & 92.0 & 21.4 \\
        4-5 edges/node & 99.2 & 99.0 & 97.8 & 99.2 & 95.6 & 94.8 & 98.8 & 94.8 & 95.9 & 96.0 & 94.9 \\
        \midrule
        Small-world & 97.2 & 96.8 & 90.1 & 91.2 & 57.0 & 48.2 & 90.5 & 83.7 & 72.1 & 69.7 & 5.9 \\
        Scale free  & 92.3 & 91.6 & 95.9 & 94.0 & 34.0 & 32.5 & 79.9 & 93.3 & 85.6 & 81.6 & 0.3 \\
        \bottomrule
    \end{tabular}
    \caption{\small \textbf{Out-of-domain generalization for quantitative experiments.} Quantitative models do not generalize to graphs of different sizes, but generalize to different number of edges, or graph structures. Graphs generated using rejection sampling seem to generalize less than those using redemption. $M_{\text{flt}}$: rejection sampling, $M_{\text{4}}$: four significant digits. }
    \label{tab:ood_quanti_results}
\end{table}

\begin{table}[h]
    \small
    \centering
    \begin{tabular}{lccccccccc}
        \toprule
         & $S_1$  &  $S_2$ & $M_1$ & $M_2$ & $M_{\text{filt}1}$ & $M_{\text{filt}2}$ & $M_{\text{4},1}$ & $M_{\text{4},2}$ & $L$ \\
        \midrule
        8-32 nodes & 99.8 & 99.9 & 4.5 & 4.5 & 4.0 & 3.9 & 4.5 & 4.5 & 0 \\
        24-32 nodes & 99.8 & 99.6 & 11.2 & 11.2 & 10.3 & 10.0 & 11.2 & 11.2 & 0 \\
        32-40 nodes  & 11.6 & 11.6 & 99.7 & 99.8 & 89.6 & 87.3 & 99.7 & 99.5 & 0.7 \\
        32-64 nodes  & 2.9 & 2.9 & 99.8 & 99.7 & 89.2 & 86.5 & 99.7 & 99.4 & 22.8 \\
        56-64 nodes  & 0 & 0 & 99.8  & 99.7 & 89.3 & 86.0 & 99.7 & 99.5 & 54.0 \\
        64-72 nodes  & 0 & 0 & 10.9 & 10.9 & 9.7 & 9.4 & 11.0 & 10.9 & 99.0 \\
        64-128 nodes  & 0 & 0 & 1.7 & 1.7  & 1.5 & 1.4 & 1.7 & 1.7 & 99.0 \\
        128-150 nodes & 0 & 0 & 0 & 0 & 0 & 0 & 0 & 0 & 4.3 \\
        \midrule
        1-2 edges/node & 98.7 & 98.9 & 97.7 & 97.5 & 52.9 & 42.9 & 98.0 & 96.7 & 90.2 \\
        4-5 edges/node & 94.0 & 94.9  & 92.8 & 92.3 & 91.9 & 91.5 & 94.3 & 94.8 & 88.0 \\
        \midrule
        Small-world & 98.6 & 98.3 & 92.6 & 92.6 & 74.5 & 60.2 & 92.8 & 91.0 & 68.3 \\
        Scale free & 99.0 & 99.3 & 96.6 & 97.0 & 46.8 & 32.7 & 97.6 & 90.7 & 88.3 \\
        \bottomrule
    \end{tabular}
    \caption{\small \textbf{Out-of-domain generalization for weighted experiments.} Quantitative models do not generalize to graphs of different sizes, but generalize to different number of edges, or graph structures. Graphs generated using rejection sampling seem to generalize less than those using redemption. $M_{\text{filt}}$: rejection sampling, $M_{\text{4}}$: four significant digits. }
    \label{tab:ood_weighted_results}
\end{table}

\clearpage

\section{The 29 metabolic networks used for testing \label{annex:kegg_networks}}

\begin{itemize}
\item Mucin type O-glycan biosynthesis
\item D-Glutamine and D-glutamate metabolism
\item Mannose type O-glycan biosynthesis
\item Lysine degradation
\item Biosynthesis of unsaturated fatty acids
\item Glycosphingolipid biosynthesis - ganglio series
\item Sphingolipid metabolism
\item Nicotinate and nicotinamide metabolism
\item Glycosylphosphatidylinositol (GPI)-anchor biosynthesis
\item Glycosphingolipid biosynthesis - lacto and neolacto series
\item Retinol metabolism
\item Phenylalanine, tyrosine and tryptophan biosynthesis
\item Tryptophan metabolism
\item alpha-Linolenic acid metabolism
\item Lipoic acid metabolism
\item Taurine and hypotaurine metabolism
\item Histidine metabolism
\item Linoleic acid metabolism
\item Caffeine metabolism
\item Glycosaminoglycan biosynthesis - chondroitin sulfate / dermatan sulfate
\item Terpenoid backbone biosynthesis
\item Primary bile acid biosynthesis
\item Phosphonate and phosphinate metabolism
\item Glycosaminoglycan biosynthesis - heparan sulfate / heparin
\item Nitrogen metabolism
\item Valine, leucine and isoleucine biosynthesis
\item Reverse Cholesterol Transport
\item Tricarboxilic Acid cycle in an engineered Synechocystis
\item A PBPK model of oral medication diffusion
\end{itemize}

%
%
%





\end{document}